\documentclass[sigconf]{acmart}

\usepackage{booktabs} 
\usepackage{amssymb}
\setcopyright{rightsretained}
\usepackage{graphicx}
\usepackage{wrapfig}
\usepackage[export]{adjustbox}
\usepackage{enumitem}

\acmArticle{4}
\acmPrice{15.00}

\begin{document}

\copyrightyear{2019}
\acmYear{2019}
\setcopyright{acmcopyright}
\acmConference[SAC '19]{The 34th ACM/SIGAPP Symposium on Applied
Computing}{April 8--12, 2019}{Limassol, Cyprus}
\acmBooktitle{The 34th ACM/SIGAPP Symposium on Applied Computing (SAC '19),
April 8--12, 2019, Limassol, Cyprus}
\acmPrice{15.00}
\acmDOI{10.1145/3297280.3297378}
\acmISBN{978-1-4503-5933-7/19/04}

\title{Few-shot classification in Named Entity Recognition Task}

\author{Alexander Fritzler}
\authornote{Neural Networks and Deep Learning Lab Moscow Institute of Physics and Technology}
\authornote{Quantum Brains}
\authornote{Higher School of Economics}
\affiliation{%
  \country{Russia}
}
\email{afritzler449@gmail.com}

\author{Varvara Logacheva}
\authornotemark[1]
\affiliation{%
  \country{Russia}
}
\email{varvara.logacheva@gmail.com}

\author{Maksim Kretov}
 \authornotemark[1]
\affiliation{%
  \country{Russia}
}
\email{kretov.mk@mipt.ru}

\begin{abstract}
For many natural language processing (NLP) tasks the amount of annotated data is limited. This urges a need to apply semi-supervised learning techniques, such as transfer learning or meta-learning. 
In this work we tackle Named Entity Recognition (NER) task using Prototypical Network --- a metric learning technique. It learns intermediate representations of words which cluster well into named entity classes. This property of the model allows classifying words with extremely limited number of training examples, and can potentially be used as a zero-shot learning method. By coupling this technique with transfer learning we achieve well-performing classifiers trained on only 20 instances of a target class.

\end{abstract}

%
%



\maketitle
\sloppy

\newcommand{\fixme}[1]{{\bf \color{red} [*FIXME* }{\em #1}{\bf ]}}

\section{Introduction}

Named Entity Recognition (NER) is the task of finding entities, such as names of persons, organizations, locations, etc. in unstructured text. These names can be individual words or phrases in a sentence. Therefore, NER is usually interpreted as sequence labelling task. This task is actively used in various information extraction frameworks and is one of the core components of goal-oriented dialogue systems~\cite{dialogues_ner}. 

When large labelled datasets are available, the task of NER can be solved with very high quality~\cite{ner_bilstm_sota}. Common benchmarks for testing new NER methods are CoNLL-2003~\cite{conll2003} and Ontonotes~\cite{ontonotes} datasets. They both include enough data to train neural architectures in a supervised learning setting.
However, in real-world applications such abundant datasets are usually not available, especially for low-resourced languages. And even if we have a large labelled corpus, it will inevitably have rare entities that occur not enough times to train a neural network to accurately identify them in text. 

This urges the need for developing methods of \textbf{few-shot} NER --- successful identification of entities for which we have extremely small number of labelled examples. One solution would be semi-supervised learning methods, i.e. methods that can yield well-performing models by combining the information from a small set of labelled data and large amounts of unlabelled data which are available for virtually any language. Word embeddings which are trained in the unsupervised manner and are used in the majority of NLP tasks as the input to a neural network, can be considered as incorporation of unlabelled data. However, they only provide general (and not always suitable) information about word meaning, whereas we argue that unsupervised data can be used to extract more task-specific information on the structure of the data.

A prominent approach to the task of learning from few examples is metric learning~\cite{metric-learning}. This term denotes techniques that learn a metric to measure fitness of an object to some class. Metric learning methods, such as matching networks~\cite{matching} and prototypical networks~\cite{prototypical}, showed good results in few-shot learning for image classification. These methods can also be considered as semi-supervised learning methods, because they use the information about structure of common objects in order to label the uncommon ones even without seeing many examples. This approach can even be used as zero-shot learning, i.e. instances of a target class do not need to be presented at training time. Therefore, such model does not need to be re-trained in order to handle new classes. This property is extremely appealing for real-world tasks.

Despite its success in image processing, metric learning has not been widely used in NLP tasks. There, in low-resourced settings researchers more often resort to transfer learning --- use of knowledge from a different domain or language. We apply prototypical networks to the NER task and compare it to commonly used baselines. We test a metric learning technique in a task which often emerges in real-world setting --- identification of instances with extremely small number of labelled examples. We show that although prototypical networks do not succeed in zero-shot NER task, they outperform other models in few-shot case.

The main contributions of the work are the following: 

\begin{enumerate}
\item we formulate few-shot NER task as a semi-supervised learning task,
\item we modify prototypical network model to enable it to solve NER task, we show that it outperforms a state-of-the-art model in low-resource setting.
\end{enumerate}

The paper is organized as follows. In Section \ref{sec:related_work} we review the existing approaches to few-shot NER task. In Section \ref{sec:prototypical} we describe the prototypical network model and its adaptation to the NER task. Section \ref{sec:few_shot_ner} defines the task and describes the models that we tested to solve it. Section \ref{sec:experimental_setup} contains the description of our experimental setup. We report and analyze our results in Section \ref{sec:results}, and in Section \ref{sec:conclusions} we conclude and provide the directions for future work.

\section{Related work}
\label{sec:related_work}

NER is a well-established task that has been solved in a variety of ways. Nowadays, as in the majority of other NLP tasks, the state of the art is sequence labelling with Recurrent Neural Networks \cite{ner_bilstm_sota,ner_sota}. However, neural architectures are very sensitive to the size of training data and tend to overfit on small datasets. Hence, the latest research on named entities concentrates on handling low-resourced cases, which often occur in narrow domains or low-resourced languages.

The work by Wang et al.~\cite{transfermed} describes feature transform between domains which allows exploiting a large out-of-domain dataset for NER task. Numerous works describe a similar transition between languages: Dandapat and Way~\cite{ner_embedding_transfer} draw correspondences between entities in different languages using a machine translation system,  Xie et al.~\cite{crosslang} map words of two languages into a shared vector space. Both these methods allow ``translating'' a big dataset to a new language. Cotterell and Duh~\cite{ner_crosslang_joint} describe a setting where the performance of a NER model for a low-resourced language is improved by training it jointly with a NER model for its well-resourced cognate. 

Besides labelled data of a different domain or language, other sources such as ontologies, knowledge bases or heuristics can be used in limited data settings~\cite{ner_ontologies}. Similarly, Tsai and Salakhutdinov~\cite{fusing} improve the image classification accuracy using side information.

Active learning is also a popular choice to reduce the amount of training data. In~\cite{activelearning} the authors apply active learning to few-shot NER task and succeed in improving the performance despite the fact that neural architectures usually require large number of training examples. A somewhat similar approach is self-learning --- training on examples labelled by a model itself. While it is ineffective in many settings,~\cite{self_learn} shows that it can improve results of few-shot NER task when combined with reinforcement learning.

The most closely related work to ours is research by Ma et al.~\cite{fine-grained} where authors learn embeddings for fine-grained NER task with hierarchical labels. They train a model to map hand-crafted and other features of words to embeddings and use mutual information metric to choose a prototype from sets of words. Analogously to this work, we aim at improving performance of NER models on rare classes. However, we do not limit the model to hierarchical classes. It makes our model more flexible and applicable to ``cold start'' problem (problem of extending data with new classes). 

Beyond NLP, there also exist multiple approaches to few-shot learning. The already mentioned metric learning technique~\cite{metric-learning} benefits from structure shared by all objects in a task, and creates a representation that shows their differences relevant to the task. Meta-learning~\cite{few_shot_meta} approach operates at two levels: it learns to solve a task from a small number of examples, and at the top level it learns more general regularities about the data across tasks. 
In~\cite{few_shot_memory} the authors demonstrate that memory-augmented neural networks, such as Neural Turing Machines, have a capacity to perform meta-learning with few labelled examples.

To the best of our knowledge, prototypical networks~\cite{prototypical} have not been applied to any NLP tasks before. They have a very attractive capacity of introducing new labels to a model without its retraining. None of models described above can perform such zero-shot learning. Although natural language is indeed different from images for which prototypical networks were originally suggested, we decided to test this model on an NLP task to see if it is possible to transfer this property to the text domain.

\section{Prototypical Networks}
\label{sec:prototypical}

\subsection{Model}
\label{section:model_theory}

Work by Snell et al. ~\cite{prototypical} introduces \textit{prototypical network} --- a model that was developed for classification in settings where labelled examples are scarce. 
This network is trained so that representations of objects returned by its last but one layer are similar for objects that belong to the same class and diverse for objects of different classes. In other words, this network maps objects to a vector space which allows easy separation of objects into meaningful task-specific clusters. 
This feature allows assigning a class to an unseen object even if the number of labelled examples of this class is very limited. 

The model is trained on two sets of examples: \textit{support set} and \textit{query set}. Support set consists of $N$ labelled examples: $S$ = \{$(\textbf{x}_1, y_1)$, ...,$(\textbf{x}_N, y_N)$\}, where each $ x_i \in \mathbb{R}^{D} $ is a $D$-dimensional representation of an object and $ y_i \in \{1,2, ..., K\} $ is the label of this object. Query set contains $N'$ labelled objects: $Q$ = \{$(\textbf{x}_1, y_1)$, ...,$(\textbf{x}_{N'}, y_{N'})$\}. Note that this partition is not stable across training steps --- the support and query sets are sampled randomly from the training data at each step.

The training is conducted in two stages:

\begin{enumerate}
	\item For each class $k$ we define $ S_k $ --- the set of objects from $S$ that belong this class. We use these sets to compute \textit{prototypes}:

$$ \textbf{c}_k = \frac{1}{\|S_k\|} \sum_i f_{\theta}(\textbf{x}_i), $$

where function $f_{\theta}: \mathbb{R}^D \to \mathbb{R}^M$ maps the input objects to the $M$-dimensional space which is supposed to keep distances between classes. $f_{\theta}$ is usually implemented as a neural network. Its architecture depends on the properties of objects.

Prototype is the averaged representation of objects in a particular class, or the centre of a cluster corresponding to this class in the $M$-dimensional space.

\item We classify objects from $Q$. In order to classify an unseen example \textbf{x}, we map it to the $M$-dimensional space using $f_{\theta}$ and then assign it to a class whose prototype is closer to the representation of \textbf{x}. We compute distance $d(f_{\theta}(\textbf{x}), \textbf{c}_k)$ for every $k$. We denote the measure of similarity of \textbf{x} to $k$ as $l_i = -d(f_{\theta}(\textbf{x}), \textbf{c}_k)$. Finally, we convert these similarities to distribution over classes using $softmax$ function: $softmax(l_1, ..., l_K)$. The model is agnostic about the distance function. Following ~\cite{prototypical}, we use squared Euclidian distance.

\end{enumerate}

The model is trained by optimising cross-entropy loss: 

$$ L(\textbf{y}, \hat{\textbf{y}}) = - \sum_{i=1}^{N'} y_i \hspace{1mm} log \hspace{1mm} \hat{y}_i, $$

where $\hat{y}_i = softmax(l_1, ..., l_K)$.


\subsection{Adaptation to NER}
In order to apply prototypical networks to NER task, we made the following changes to the baseline model described above:

\paragraph{\textbf{Sequential vs independent objects}} Image dataset contains separate images that are not related to each other. 
In contrast, in NLP tasks we often need to classify words which are grouped in sequences. Words in a sentence influence each other, and when labelling a word we should take into account labels of neighbouring words. Considering a word in isolation does not make sense in such setting. Nevertheless, in NER task we need to classify separate words, so following the description of the model from the previous section, we should assemble the support set $S$ from pairs ($w_i$, $y_i$), where $w_i$ is a word and $y_i$ is its label. However, this division can break the sentence structure, if some words in a sentence are assigned to the support set and others to query set. In order to prevent such situations we form our support and query sets from whole sentences.

\paragraph{\textbf{Class ``no entity''}} In NER task we have class \textit{O} that is used to denote words which are not named entities. It cannot be interpreted in the same way as other classes, because objects of class \textit{O} do not need to (and should not) be close to each other in a vector space. In order to mitigate this problem we modified our prediction function $softmax(l_1, ..., l_K)$. We replaced the similarity score $l_O$ for the \textit{O} class with a scalar $b_{O}$, and used the following form of softmax: $softmax(l_1, ...,  l_{K-1}, b_{O})$. $b_O$ is trained along with parameters $\theta$ of the model. The initial value of $b_O$ is a hyper-parameter.

\paragraph{\textbf{In-domain and out-of-domain training}} 
In original paper describing prototypical networks~\cite{prototypical} they were applied to the setting of zero-shot learning. Weights of the model are updated during training phase, but once training is over instances from test classes are only used for calculation of prototypes. Given it is usually easy to obtain few labelled examples, we modified original \textit{zero-shot} setting to \textit{few-shot} setting: we use a small number of available labelled examples of the target class during training phase. We denote this data as \textbf{in-domain} training set, and data for other classes is referred to as \textbf{out-of-domain} training. Here \textit{domains} in the traditional NLP sense are the same --- texts come from the same sources and word distributions are similar. Here we refer to discrepancy between sets of named entity classes that they use.


\section{Few-shot NER}
\label{sec:few_shot_ner}

\subsection{Task formulation}

NER is a sequence labelling task, where each word in a sentence is assigned either one of entity classes (``Person'', ``Location'', ``Organisation'', etc.) or \textit{O} class if it is not one of the desired entities. 

While common classes are usually identified correctly by the existing methods, we target particularly at rare classes for which we have only a very limited number of labelled examples. To increase the quality of their identification, we use the information from other classes. Therefore, we train a separate model for every class in order to see the performance on each of them in isolation. Such formulation can also be considered as a way to tackle the ``cold start'' problem --- adapting a NER model to label entities of a new class with very little number of entities. 

As it was described above, we have two training sets: \textit{out-of-domain} and \textit{in-domain}. Since we simulate the ``cold start'' problem in our experiments, these datasets have the following characteristics. The \textit{out-of-domain} data is quite large and labelled with a number of named entity classes except the target class $C$ --- this is the initially available data. The \textit{in-domain} dataset is very small and contains labels only for the class $C$ --- this is the new data which we acquire afterwards and which we would like to infuse into the model.

In order to train a realistic model we need to keep the frequency of $C$ in our \textit{in-domain} training data similar to the frequency of this class in general distribution. Therefore, if instances of this class occur on average in one of three sentences, then our \textit{in-domain} training data has to contain sentences with no instances of class $C$ (``empty'' sentences), and their number should be twice as larger as the number of sentences with $C$. In practice this can be achieved by sampling sentences from unlabelled data until we obtain the needed number of instances of class $C$. 

\subsection{Basic models}
\label{sec:basic_models}

We use two main architectures --- the commonly used RNN baseline and a prototypical network adapted for the NER task. Other models we test use these two models as building blocks.

\paragraph{\textbf{RNN + CRF model}}

As our baseline we use a NER model implemented in AllenNLP open-source library~\cite{allen}. The model processes sentences in the following way:

\begin{enumerate}
	\item words are mapped to pre-trained embeddings (any embeddings, such as GloVe~\cite{glove}, ELMo~\cite{elmo}, etc. can be used)
    \item additional word embedding are produced using a character-level trainable Recurrent Neural Network (RNN) with LSTM cells,
    \item embeddings produced at stages (1) and (2) are concatenated and used as the input to a bi-directional RNN with LSTM cells. This network processes the whole sentence and creates context-dependent representations of every word
    \item a feed-forward layer converts hidden states of the RNN from stage (3) to logits that correspond to every label,
    \item the logits are used as input to a Conditional Random Field (CRF)~\cite{crf} model that outputs the probability distribution of tags for every word in a sentence.
\end{enumerate}

The model is trained by minimizing negative log-likelihood of true tag sequences. It has to be noted that this baseline is quite reasonable even in our limited resource setting. 

\paragraph{\textbf{Prototypical Network}}

\begin{figure}[h]
\includegraphics[width=1.0\linewidth]{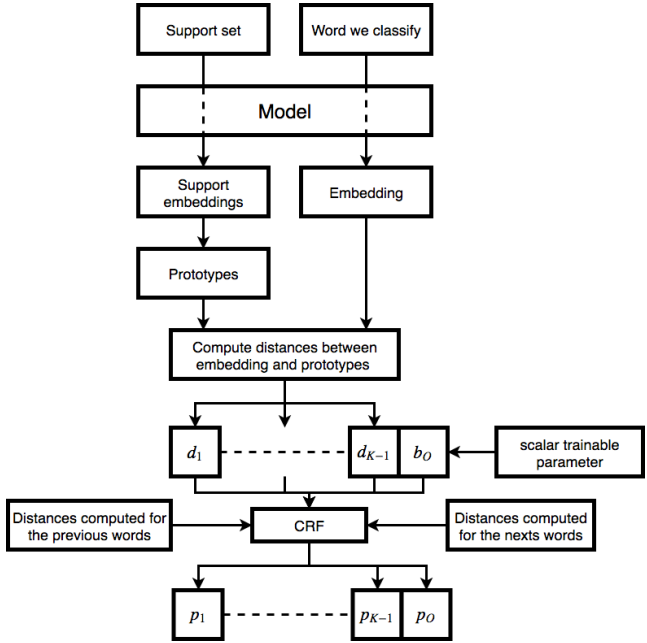} 
\caption{Model architecture}
\label{fig:subim1}
\end{figure}

The architecture of the prototypical network that we use for NER task is very similar to the one of our baseline model. The main change concerns the feed-forward layer. While in the baseline model it transforms RNN hidden states to logits corresponding to labels, in our prototypical network it maps these hidden states to the $M$-dimensional space. The output of the feed-forward layer is then used to construct prototypes from the support set. These prototypes are used to classify examples from the query set as described in section \ref{section:model_theory}.
We try variants of this model both with and without the CRF layer. 
The architecture of the prototypical network model is provided in Figure \ref{fig:subim1}.

\subsection{Experiments}

We perform experiments with a number of different models. We test the different variants of prototypical network model and compare them with RNN baseline. In addition to that, we try transfer learning scenario and combine it with these models. Here we provide the description of all models we test.

\paragraph{\textbf{RNN Baseline (Base)}}

This is the baseline RNN model described above. We train it using only \textit{in-domain} training set.

\paragraph{\textbf{Baseline prototypical network (BaseProto)}}

This is the baseline prototypical network model. We train it on \textit{in-domain} training data. We divide it into two parts. If the \textit{in-domain} set contains $N$ sentences with instances of the target class $C$ and $V$ sentences ``empty'' sentences, we use $N/2$ sentences with instances of $C$ as support set, and other $N/2$ such sentences along with $V/2$ ``empty'' sentences serve as query set. We use only half of ``empty'' sentences to keep the original frequency of class $C$ in the query set. Note that the partition is new for every training iteration.

\paragraph{\textbf{Regularised prototypical network (Protonet)}}

The architecture and training procedure of this model are the same as those of \textit{BaseProto} model. The only difference is the data we use for training. At each training step we select the training data using one of two scenarios:

\begin{enumerate}
    \item we use \textit{in-domain} training data, i.e. data labelled with the target class $C$ (this setup is the same as the one we use in \textit{BaseProto}),
    \item we change the target class: we (i) randomly select a new target class $C'$ ($C' \neq C$), (ii) sample sentences from \textit{out-of-domain} dataset until we find $N$ instances of $C'$, and (iii) re-label the sampled sentences so that they contain only labels of class $C'$.
\end{enumerate}

At each step we choose the scenario (1) with probability $p$, or scenario (2) with probability $(1-p)$.

Therefore, throughout training the network is trained to predict our target class (scenario (1)), but occasionally it sees instances of some other classes and constructs prototypes for them (scenario (2)). We suggest that this model can be more efficient than BaseProto, because at training time it is exposed to objects of different classes, and the procedure that maps objects to prototype space becomes more robust. This is also a way to leverage out-of-domain training data.

\paragraph{\textbf{Transfer learning baseline (WarmBase)}}

We test a common transfer learning model --- use of knowledge about out-of-domain data to label in-domain samples. The training of this model is two-part:

\begin{enumerate}
	\item We train our baseline RNN+CRF model using \textit{out-of-domain} training set.
    \item We save all weights of the model except CRF and label prediction layer, and train this model again using \textit{in-domain} training set.
\end{enumerate}

\paragraph{\textbf{Transfer learning + prototypical network (WarmProto)}}

In addition to that, we combine prototypical network with pre-training on out-of-domain data. We first train a Base model on the \textit{out-of-domain} training set. Then, we train a Protonet model as described above, but initialise its weights with weights of this pre-trained Base model.

\paragraph{\textbf{WarmProto-CRF}}

This is the same prototypical network pre-trained on \textit{out-of-domain} data, but it is extended with a CRF layer on top of logits as described in section \ref{sec:basic_models}.

\paragraph{\textbf{WarmProto for zero-shot training (WarmProtoZero)}}

We train the same WarmProto model, but with the probability $p$ set to 0. In other words, our model does not see instances of the target class at training time. It learns to produce representations on objects of other classes. Then, at test time, it is given $N$ entities of the target class as support set, and words in test sentences are assigned to either this class or \textit{O} class based on their similarity to this prototype. This is the only zero-shot learning scenario that we test.

\section{Experimental setup}
\label{sec:experimental_setup}

\subsection{Dataset}

\begin{figure}
	\centering
	\includegraphics[width=0.9\linewidth]{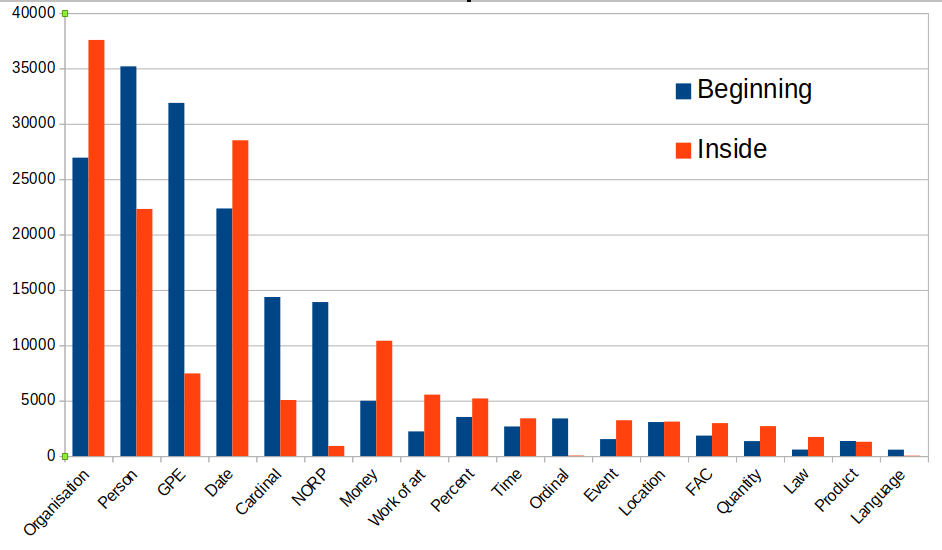}
	\caption{Ontonotes dataset --- statistics of classes frequency in the training data.}
	\label{fig:ontonotes_stat}
\end{figure}

We conduct all our experiments on the Ontonotes dataset~\cite{ontonotes}. It contains 18 classes (+ $O$ class). The classes are not evenly distributed --- the training set contains over $30.000$ instances of some common classes and less than 100 for rare classes. The distribution of classes is shown in Figure \ref{fig:ontonotes_stat}. The size of the training set is $150.374$ sentences, the size of the validation set is $19.206$ sentences.

As the majority of NER datasets, Ontonotes adopts \textbf{BIO} (\textbf{B}eginning, \textbf{I}nside, and \textbf{O}utside) labelling. It provides an extension to class labels, namely, all class labels except \textit{O} are prepended with the symbol ``B'' if the corresponding word is the first (or the only) word in an entity, or with the symbol ``I'' otherwise.

\subsection{Data preparation: simulated few-shot experiments}

We use the Ontonotes training data as \textit{out-of-domain} training set (where applicable) and sample \textit{in-domain} examples from the validation set. In our formulation, the \textit{in-domain} data is the data where only instances of a target class $C$ (class we want to predict) are labelled. Conversely, the \textit{out-of-domain} data contains instances of some set of classes, but not of the target class. Therefore, we prepare our data by replacing all labels \textit{B-C} and \textit{I-C} with \textit{O} in the training data, and in the validation data we replace all labels \textit{except} \textit{B-C} and \textit{I-C} with \textit{O}. Note that since we run the experiments for each of 18 Ontonotes classes, we perform this re-labelling for every experiment.

The validation data is still too large for our low-resourced scenario, so we use only a part of it for training. We sample our \textit{in-domain} training data as follows. We randomly select sentences from the re-labelled validation set until we obtain $N$ sentences with at least one instance of the class $C$. Note that sentences of the validation set are not guaranteed to have instances of $C$, so our training data can have some ``empty'' sentences, i.e. sentences where all words are labelled with \textit{O}. This sampling procedure allows keeping the proportion of instances of class \textit{C} close to the one of the general distribution.

In our preliminary experiments we noticed that such sampling procedure leads to large variation in the final scores, because the size of \textit{in-domain} training data can vary significantly. In order to reduce this variation we alter the sampling procedure. We define a function $pr(C)$ which computes the proportion of labels of a class $C$ in the validation set ($pr(C) \in [0, 1]$). Then we sample $N$ sentences containing instances of class $C$ and $\frac{N \times (1-pr(C))}{pr(C)}$ sentences without class $C$. Thus, we keep the proportion instances of class $C$ in our \textit{in-domain} training dataset equal to that of the validation set. We use the same procedure when sampling training examples from \textit{out-of-domain} data for \textit{Protonet} model.

\subsection{Design of experiments}

We conduct separate experiments for each of 18 Ontonotes classes. For each class we conducted 4 experiments with different random seeds. We report averaged results for each class.

We design separate experiments for selection of hyper-parameters and the optimal number of training epochs. For that we selected three well-represented classes --- ``GPE'' (geopolitical entity), ``Date'', and ``Org'' (organization) --- to conduct \textit{validation} experiments on them. We selected training sets as described above, and used the test set (consisting of $\approx 19.000$ sentences) to tune hyper-parameters and to stop the training. For other classes we did not perform hyper-parameter tuning. Instead, we used the values acquired in the validation experiments with the three validation classes. In these experiments we used the test set only for computing the performance of trained models.

\begin{figure*}[ht!]
\includegraphics[scale=0.5]{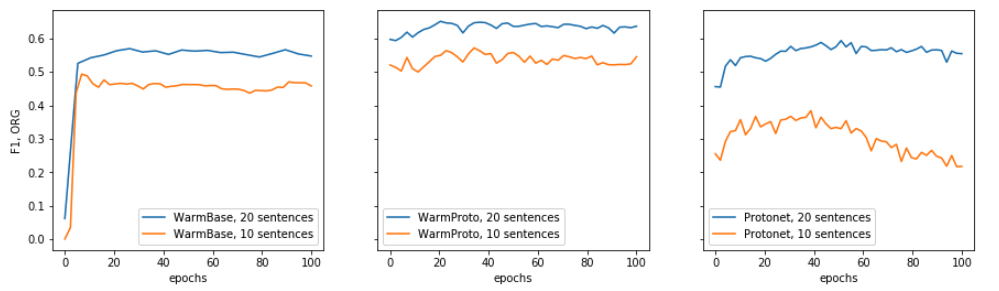} 
\caption{Performance of models trained on 10 and 20 sentences.}
\label{fig:10_vs_20}
\end{figure*}

The motivation of such setup is the following. In many few-shot scenarios researchers report experiments where they train on a small training set, and tune the model on a very large validation set. We argue that this scenario is unrealistic, because if we had a large number of labelled examples in a real-world problem, it would be more efficient to use them for training, and not for validation. On the other hand, a more realistic scenario is to have a very limited number of labelled sentences overall. In that case we could still reserve a part of them for validation. However, we argue that this is also inefficient. If we have 20 examples and decide to train on 10 of them and validate on another 10, this validation will be inaccurate, because 10 examples are not enough to evaluate the performance of a model reliably. Therefore, our evaluation will be very noisy and is likely to result in sub-optimal values of hyper-parameters. On the other hand, additional 10 examples can boost the quality of a model, as it can be seen in Figure \ref{fig:10_vs_20}. Therefore, we assume that optimal hyperparameters are the same for all labels, and use the values we found in validation experiments.

\begin{table*}[ht!]
\begin{center}
\begin{tabular}{|l|ccccccc|}
\hline \bf Class name & \bf Base & \bf BaseProto  & \bf WarmProtoZero & \bf Protonet & \bf WarmProto  & \bf WarmBase & \bf WarmProto-CRF\\
\hline
 \multicolumn{8}{|c|}{\textbf{Validation Classes}} \\
\hline
GPE & 69.75 $\pm$ 9.04 & 69.8 $\pm$ 4.16 & 60.1 $\pm$ 5.56 & 78.4 $\pm$ 1.19 & \textbf{83.62} $\pm$ \textbf{3.89} & 75.8 $\pm$ 6.2 & \underline{80.05} $\pm$ \underline{5.4} \\
DATE & 54.42 $\pm$ 3.64 & 50.75 $\pm$ 5.38 & 11.23 $\pm$ 4.57 & 56.55 $\pm$ 4.2 & \underline{61.68} $\pm$ \underline{3.38} & 56.32 $\pm$ 2.32 & \textbf{65.42} $\pm$ \textbf{2.82} \\
ORG & 42.7 $\pm$ 5.54 & 39.1 $\pm$ 7.5 & 17.18 $\pm$ 3.77 & 56.35 $\pm$ 2.86 & \underline{63.75} $\pm$ \underline{2.43} & 63.45 $\pm$ 1.79 & \textbf{69.2} $\pm$ \textbf{1.2} \\
\hline
 \multicolumn{8}{|c|}{\textbf{Test Classes}} \\
\hline
EVENT & 32.33 $\pm$ 4.38 & 24.15 $\pm$ 4.38 & 4.85 $\pm$ 1.88 & 33.95 $\pm$ 5.68 & 33.85 $\pm$ 5.91 & \underline{35.15} $\pm$ \underline{4.04} & \textbf{45.2} $\pm$ \textbf{4.4} \\
LOC & 31.75 $\pm$ 9.68 & 24.0 $\pm$ 5.56 & 16.62 $\pm$ 7.18 & 42.88 $\pm$ 2.03 & \underline{49.1} $\pm$ \underline{2.4} & 40.67 $\pm$ 4.85 & \textbf{52.0} $\pm$ \textbf{4.34} \\
FAC & 36.7 $\pm$ 8.15 & 29.83 $\pm$ 5.58 & 6.93 $\pm$ 0.62 & 41.05 $\pm$ 2.74 & \underline{49.88} $\pm$ \underline{3.39} & 45.4 $\pm$ 3.01 & \textbf{56.85} $\pm$ \textbf{1.52} \\
CARDINAL & 54.82 $\pm$ 1.87 & 53.7 $\pm$ 4.81 & 8.12 $\pm$ 7.92 & 64.05 $\pm$ 1.61 & \underline{66.12} $\pm$ \underline{0.43} & 62.98 $\pm$ 3.5 & \textbf{70.43} $\pm$ \textbf{3.43} \\
QUANTITY & 64.3 $\pm$ 5.06 & 61.72 $\pm$ 4.9 & 12.88 $\pm$ 4.13 & 65.05 $\pm$ 8.64 & 67.07 $\pm$ 5.11 & \underline{69.65} $\pm$ \underline{5.8} & \textbf{76.35} $\pm$ \textbf{3.09} \\
NORP & 73.5 $\pm$ 2.3 & 72.1 $\pm$ 6.0 & 39.92 $\pm$ 10.5 & \underline{83.02} $\pm$ \underline{1.42} & \textbf{84.52} $\pm$ \textbf{2.79} & 79.53 $\pm$ 1.32 & 82.4 $\pm$ 1.15 \\
ORDINAL & 68.97 $\pm$ 6.16 & 71.65 $\pm$ 3.31 & 1.93 $\pm$ 3.25 & \textbf{76.08} $\pm$ \textbf{3.55} & 73.05 $\pm$ 7.14 & 69.77 $\pm$ 4.97 & \underline{75.52} $\pm$ \underline{5.11} \\
WORK\_OF\_ART & \underline{30.48} $\pm$ \underline{1.42} & 27.5 $\pm$ 2.93 & 3.4 $\pm$ 2.37 & 28.0 $\pm$ 3.33 & 23.48 $\pm$ 5.02 & 30.2 $\pm$ 1.27 & \textbf{32.25} $\pm$ \textbf{3.11} \\
PERSON & 70.05 $\pm$ 6.7 & 74.1 $\pm$ 5.32 & 38.88 $\pm$ 7.64 & \underline{80.53} $\pm$ \underline{2.15} & 80.42 $\pm$ 2.13 & 78.03 $\pm$ 3.98 & \textbf{82.32} $\pm$ \textbf{2.51} \\
LANGUAGE & \underline{72.4} $\pm$ \underline{5.53} & 70.78 $\pm$ 2.62 & 4.25 $\pm$ 0.42 & 68.75 $\pm$ 6.36 & 48.77 $\pm$ 17.42 & 65.92 $\pm$ 3.52 & \textbf{75.62} $\pm$ \textbf{7.22} \\
LAW & \underline{58.08} $\pm$ \underline{4.9} & 53.12 $\pm$ 4.54 & 2.4 $\pm$ 1.15 & 48.38 $\pm$ 8.0 & 50.15 $\pm$ 7.56 & \textbf{60.13} $\pm$ \textbf{6.08} & 57.72 $\pm$ 7.06 \\
MONEY & 70.12 $\pm$ 5.19 & 66.05 $\pm$ 1.66 & 12.48 $\pm$ 11.92 & 68.4 $\pm$ 6.3 & \underline{73.68} $\pm$ \underline{4.72} & 68.4 $\pm$ 5.08 & \textbf{79.35} $\pm$ \textbf{3.6} \\
PERCENT & 76.88 $\pm$ 2.93 & 75.55 $\pm$ 4.17 & 1.82 $\pm$ 1.81 & 80.18 $\pm$ 4.81 & \underline{85.3} $\pm$ \underline{3.68} & 79.2 $\pm$ 3.76 & \textbf{88.32} $\pm$ \textbf{2.76} \\
PRODUCT & 43.6 $\pm$ 7.21 & \underline{44.35} $\pm$ \underline{3.48} & 3.75 $\pm$ 0.58 & 39.92 $\pm$ 7.22 & 35.1 $\pm$ 9.35 & 43.4 $\pm$ 8.43 & \textbf{49.32} $\pm$ \textbf{2.92} \\
TIME & 35.93 $\pm$ 6.35 & 35.8 $\pm$ 2.61 & 8.02 $\pm$ 3.05 & 50.15 $\pm$ 5.12 & \underline{56.6} $\pm$ \underline{2.28} & 45.62 $\pm$ 5.64 & \textbf{59.8} $\pm$ \textbf{0.76} \\
\hline
\end{tabular}
\end{center}
\caption{Results of experiments in terms of chunk-based $F_1$-score. Numbers in bold mean the best score for a particular class, underlined numbers are the second best results. Numbers are averaged across 4 runs with standard deviations calculated.}
\label{table:main_results}
\end{table*}

\subsection{Model parameters}

In all our experiments we set $N$ (number of instances of the target class in \textit{in-domain} training data) to 20. This number of examples is small enough and can be easily labelled by hand. At the same time, it produces models of reasonable quality. Figure \ref{fig:10_vs_20} compares the performance of models trained on 10 and 20 examples. We see the significant boost in performance for the latter case. Moreover, in the rightmost plot the learning curve for the smaller dataset goes down after the 40-th epoch, which does not happen when the larger dataset is used. This shows that $N=20$ is a reasonable trade-off between model performance and cost of labelling.

In the \textit{Protonet} model we set $p$ to 0.5. Therefore, the model is trained on the instances of the target class $C$ half of the steps, and another half of the times it is shown instances of some other randomly chosen class. 

We optimize all models with Adam optimizer in {\tt pytorch} implementation. Base and WarmBase methods use batches of 10 sentences during in-domain training. We train out-of-domain RNN baseline (warm-up for WarmBase and WarmProto* models) using batch of size 32. All models based on prototypical network use batches of size 100 --- 40 in support set and 60 in query set.   
 We also use L2-regularization with a multiplier 0.1. All models are evaluated in terms of chunk-based $F_1$-score for the target class \cite{conll2003}.

The open-source implementation of the models is available online.\footnote{\url{https://github.com/Fritz449/ProtoNER}}

\section{Results}
\label{sec:results}

\subsection{Performance of models}

We selected hyperparameters in the validation experiment and then used them when training models for other classes. We use the following values. The initial value of $b_O$ (logit for the \textit{O} class) is set to -4. We use dropout with rate 0.5 in LSTM cells for all our experiments. The dimensionality of embeddings space $M$ for all models based on prototypical network is set to 64. For all models we use learning rate of $3e-3$.

Table \ref{table:main_results} shows the results of our experiments for all classes and methods. It is clearly seen that 20 sentences is not enough to train a baseline RNN+CRF model. Moreover, we see that the baseline prototypical network (\textit{BaseProto}) performs closely to the RNN baseline. This shows that 20 instances of the target class is also not enough to construct a reliable prototype.

On the other hand, if a prototypical network is occasionally exposed to instances of other classes, as it is done in \textit{Protonet} model, then the prototypes it constructs are better at identifying the target class. \textit{Protonet} shows better results than \textit{Base} and \textit{BaseProto} on many classes. 

The transfer learning baseline (\textit{WarmBase}) achieves results which are comparable with those of \textit{Protonet}. This allows to conclude that the information on structure of objects of other classes is helpful even for conventional RNN baseline, and pre-training on out-of-domain data is useful.

Prototypical network pre-trained on out-of-domain data (\textit{WarmProto}) beats \textit{WarmBase} and \textit{Protonet} in more than half of experiments. Analogously to transfer learning baseline, it benefits from the use of out-of-domain data. Unfortunately, such model is not suitable for zero-shot learning --- the \textit{WarmProtoZero} model performs below any other models including the RNN baseline.

Finally, if we enable CRF layer of \textit{WarmProto} model, the performance grows sharply. As we can see, \textit{WarmProto-CRF} beats all other models in almost all experiments.
Thus, prototypical network is more effective than RNN baseline in the setting where in-domain data is extremely limited. 

\subsection{Influence of BIO labelling}

When such a small number of entities is available, the BIO labelling used in NER datasets can harm the performance of models. First of all, the majority of entities can contain only one word, and the number of \textit{I} tags can be too small if there are only 20 entities overall. This can decrease the quality of predicting these tags dramatically. Another potential problem is that words labelled with \textit{B} and \textit{I} tags can be similar, and a model can have difficulties distinguishing between them using prototypes. Again, this effect can be amplified by the fact that very small number of instances is used for training, and prototypes themselves have high variance.

\begin{table*}[ht!]
\begin{center}
\begin{tabular}{|l|cccc|}
\hline \bf Class name & \bf WarmBase + BIO & \bf WarmBase + TO & \bf WarmProto + BIO & \bf WarmProto + TO\\
\hline
 \multicolumn{5}{|c|}{\textbf{Validation Classes}} \\
\hline
GPE & 75.8 $\pm$ 6.2 & 74.8 $\pm$ 4.16 & \textbf{83.62} $\pm$ \textbf{3.89} & \underline{82.02} $\pm$ \underline{0.42} \\
DATE & 56.32 $\pm$ 2.32 & 58.02 $\pm$ 2.83 & \underline{61.68} $\pm$ \underline{3.38} & \textbf{64.68} $\pm$ \textbf{3.65} \\
ORG & 63.45 $\pm$ 1.79 & 62.17 $\pm$ 2.9 & \underline{63.75} $\pm$ \underline{2.43} & \textbf{65.22} $\pm$ \textbf{2.83} \\
\hline
 \multicolumn{5}{|c|}{\textbf{Test Classes}} \\
\hline
EVENT & \underline{35.15} $\pm$ \underline{4.04} & \textbf{35.4} $\pm$ \textbf{6.04} & 33.85 $\pm$ 5.91 & 34.75 $\pm$ 2.56 \\
LOC & 40.67 $\pm$ 4.85 & 40.08 $\pm$ 2.77 & \textbf{49.1} $\pm$ \textbf{2.4} & \underline{49.05} $\pm$ \underline{1.04} \\
FAC & \underline{45.4} $\pm$ \underline{3.01} & 44.88 $\pm$ 5.82 & \textbf{49.88} $\pm$ \textbf{3.39} & 43.52 $\pm$ 3.09 \\
CARDINAL & 62.98 $\pm$ 3.5 & 63.27 $\pm$ 3.66 & \underline{66.12} $\pm$ \underline{0.43} & \textbf{69.2} $\pm$ \textbf{1.51} \\
QUANTITY & \textbf{69.65} $\pm$ \textbf{5.8} & \underline{69.3} $\pm$ \underline{3.41} & 67.07 $\pm$ 5.11 & 67.97 $\pm$ 2.98 \\
NORP & 79.53 $\pm$ 1.32 & 80.75 $\pm$ 2.38 & \textbf{84.52} $\pm$ \textbf{2.79} & \underline{84.5} $\pm$ \underline{1.61} \\
ORDINAL & 69.77 $\pm$ 4.97 & 70.9 $\pm$ 6.34 & \underline{73.05} $\pm$ \underline{7.14} & \textbf{74.7} $\pm$ \textbf{4.94} \\
WORK\_OF\_ART & \textbf{30.2} $\pm$ \textbf{1.27} & \underline{25.78} $\pm$ \underline{4.07} & 23.48 $\pm$ 5.02 & 25.6 $\pm$ 2.86 \\
PERSON & 78.03 $\pm$ 3.98 & 76.0 $\pm$ 3.12 & \textbf{80.42} $\pm$ \textbf{2.13} & \underline{78.8} $\pm$ \underline{0.26} \\
\hline
\end{tabular}
\end{center}
\caption{$F_1$-scores for models WarmBase and WarmProto trained on data with and without BIO labelling. Numbers in bold mean the best score for a particular class, underlined numbers are the second best results. Numbers are averaged across 4 runs with standard deviations calculated.}
\label{table:bio_tagging}
\end{table*}

In order to check if these problems hamper the performance of our models, we performed another set of experiments. We removed BIO tagging --- for the target class $C$ we replaced both \textit{B-C} and \textit{I-C} with \textit{C}. This \textbf{TO} (tag/other) tagging reduced sparsity in the training data. We did so for both in-domain and out-of-domain training sets. The test set remained the same, because chunk-based $F_1$-score we use for evaluation is not affected by differences between BIO and TO labelling, it always considers a named entity as a whole.

Table \ref{table:bio_tagging} shows the result of WarmBase and WarmProto models trained on BIO-labelled and TO-labelled data. 
It turns out that in the majority of cases the differences between $F_1$-scores of these models are not significant. Therefore, BIO labelling does not affect our models.

\section{Conclusions}
\label{sec:conclusions}

In this work we suggested solving the task of NER with \textit{metric learning} technique actively used in other Machine Learning tasks but rarely applied to NLP. We adapted a metric learning method, namely, prototypical network originally used for image classification to analysis of text. It projects all objects into a vector space which keeps distances between classes, so objects of one class are mapped to similar vectors. These mappings form a \textit{prototype} of a class, and at test time we assign new objects to classes by similarity of an object representation to class prototype.

In addition to that, we considered the task of NER in a semi-supervised setting --- we identified our target classes in text using the information about words of other classes. We showed that prototypical network is more effective in such setting than the state-of-the-art RNN model. Unlike RNN, prototypical network is suitable in cases where extremely small amount of data is available.

According to the original formulation of prototypical network, it can be used as zero-shot learning method, i.e. method which can assign an object to a particular class without seeing instances of this class at training time. We experimented with zero-shot setting for NER and showed that prototypical networks can in principle be used for zero-shot text classification, although there is still much room for improvement. We suggest that this is a prominent direction of future research.

We saw that prototypical networks shows considerably different performance on different classes of named entities. It would be interesting to perform more thorough qualitative analysis to identify characteristics of textual data which is more suitable for this method. 

Finally, in our current experiments we trained models to predict entities of only a single class. In our future work we would like to check if the good performance of prototypical network scales to multiple classes. We will focus on training a prototypical network that can predict all classes of Ontonotes or another NER dataset at once.


\bibliographystyle{ACM-Reference-Format}
\bibliography{sample-bibliography} 

\end{document}